\documentclass[sigconf]{acmart}
\usepackage{bm}
\usepackage{multirow}
\usepackage{colortbl}
\usepackage{booktabs}
\usepackage{hyperref}
\usepackage{balance}
\usepackage{subfigure}
\AtBeginDocument{%
  \providecommand\BibTeX{{%
    \normalfont B\kern-0.5em{\scshape i\kern-0.25em b}\kern-0.8em\TeX}}}

\setcopyright{iw3c2w3}
\copyrightyear{2021}
\acmYear{2021}

\acmDOI{10.1145/3442381.3450042}

\acmConference[WWW '21]{Proceedings of the Web Conference 2021}{April 19--23, 2021}{Ljubljana, Slovenia}
\acmBooktitle{Proceedings of the Web Conference 2021 (WWW '21), April 19--23, 2021,
Ljubljana, Slovenia}
\acmPrice{}
\acmISBN{978-1-4503-8312-7/21/04}

\begin{document}

% \newcommand\BibTeX{B\textsc{ib}\TeX}
% \newcommand{\todo}[1]{{\color{red}#1}}
% \newcommand{\cjy}[1]{{\color{cyan}#1}}
% \newcommand{\jeff}[1]{{\color{blue}#1}}
% \newcommand{\chj}[1]{{\color{red}#1}}

% \newcommand{\todo}[1]{{\color{black}#1}}
% \newcommand{\update}[1]{{\color{black}#1}}
% \newcommand{\cjy}[1]{{\color{black}#1}}
% \newcommand{\jeff}[1]{{\color{black}#1}}
% \newcommand{\chj}[1]{{\color{black}#1}}
% \newcommand{\dsm}[1]{{\color{black}#1}}

%\title{OntoZSL: Ontology-enhanced Prior Knowledge for Zero-shot Learning and Its Applications}
\title{OntoZSL: Ontology-enhanced Zero-shot Learning}
\author{Yuxia Geng}
\email{gengyx@zju.edu.cn}
\affiliation{%
  \institution{Zhejiang University}
  \city{Hangzhou}
  \country{China}
}

\author{Jiaoyan Chen}
\email{jiaoyan.chen@cs.ox.ac.uk}
\affiliation{%
  \institution{
  University of Oxford}
  \city{Oxford}
  \country{United Kingdom}}

\author{Zhuo Chen}
\email{zhuo.chen@zju.edu.cn}
\affiliation{%
  \institution{Zhejiang University}
  \city{Hangzhou}
  \country{China}
}

\author{Jeff Z. Pan}
\email{https://knowledge-representation.org/j.z.pan/}
\affiliation{%
 \institution{University of Edinburgh}
 \city{Edinburgh}
 \country{United Kingdom}}

\author{Zhiquan Ye}
\email{yezq@zju.edu.cn}
\affiliation{%
  \institution{
 College of Computer Science, \\ 
  Zhejiang University}
  \city{Hangzhou}
  \country{China}
}

\author{Zonggang Yuan}
\email{yuanzonggang@huawei.com}
\affiliation{%
%   \institution{Huawei NAIE CTO Office}
NAIE CTO Office, \\
Huawei Technologies Co., Ltd.
  \city{Nanjing}
  \country{China}}

\author{Yantao Jia}
\email{jamaths.h@163.com}
\affiliation{
\institution{
Poisson Lab, \\
Huawei Technologies Co., Ltd.}
\city{Beijing}
\country{China}}

\author{Huajun Chen}
\authornote{Corresponding author.}
\email{huajunsir@zju.edu.cn}
\affiliation{
% \institution{Zhejiang University 
% \& AZFT Joint Lab for Knowledge Engine \& ZJU-Hangzhou Innovation Center
% }
\institution{
College of Computer Science \& HIC,
Zhejiang University \\ 
AZFT Knowledge Engine Lab
}
}

\renewcommand{\shortauthors}{Geng and Chen, et al.}

\begin{abstract}

Zero-shot Learning (ZSL), which aims to predict for those classes that have never appeared in the training data,
has arisen hot research interests.
The key of implementing ZSL is to leverage the prior knowledge of classes which builds the semantic relationship between classes and enables the transfer of the learned models (e.g., features) from training classes (i.e., seen classes) to unseen classes.
However, the priors adopted by the existing methods are relatively limited with incomplete semantics.
In this paper, we explore richer and more competitive prior knowledge to model the inter-class relationship for ZSL via ontology-based knowledge representation and semantic embedding.
Meanwhile, to address the data imbalance between seen classes and unseen classes, we developed a generative ZSL framework with Generative Adversarial Networks (GANs).

Our main findings include: (i) an ontology-enhanced ZSL framework that can be applied to different domains, such as image classification (IMGC) and knowledge graph completion (KGC);
(ii) a comprehensive evaluation with multiple zero-shot datasets from different domains, where our method often achieves better performance than the state-of-the-art models.
In particular, on four representative ZSL baselines of IMGC, the ontology-based class semantics outperform the previous priors e.g., the word embeddings of classes by an average of $12.4$ accuracy points in the standard ZSL across two example datasets (see Figure~\ref{fig:class_embed_results}).

\end{abstract}



% <concept_id>10010147.10010178.10010187.10010188</concept_id>
% <concept_desc>Computing methodologies~Semantic networks</concept_desc>

% \ccsdesc[500]{Computing methodologies~Semantic networks}



\maketitle

\section{Introduction}

Machine learning often operates on a closed world assumption: it trains the model with a number of labeled samples and makes predictions with classes that have appeared in the training stage (i.e., seen classes).
For those newly emerging classes, hundreds of samples are needed to be collected and labeled.
However, it is impractical to always annotate enough samples and retrain the model for all the emerging classes.
Targeting such a limitation, Zero-shot Learning (ZSL) was proposed to handle these novel classes without seeing their training samples (i.e., unseen classes).
Over the past few years, ZSL has been introduced in a wide range of machine learning tasks, such as image classification \cite{xian2019survey,frome2013devise,lampert2013attribute,DBLP:conf/cvpr/LiJLD0H19}, relation extraction \cite{li2020logic} and knowledge graph completion \cite{Qin2020ZSGAN,xie2016representation}.
 
\begin{figure*}[htbp]
\centering
\subfigure[Zero-shot Image Classification]{
\begin{minipage}[c]{0.8\linewidth}
\centering
\includegraphics[width=14cm]{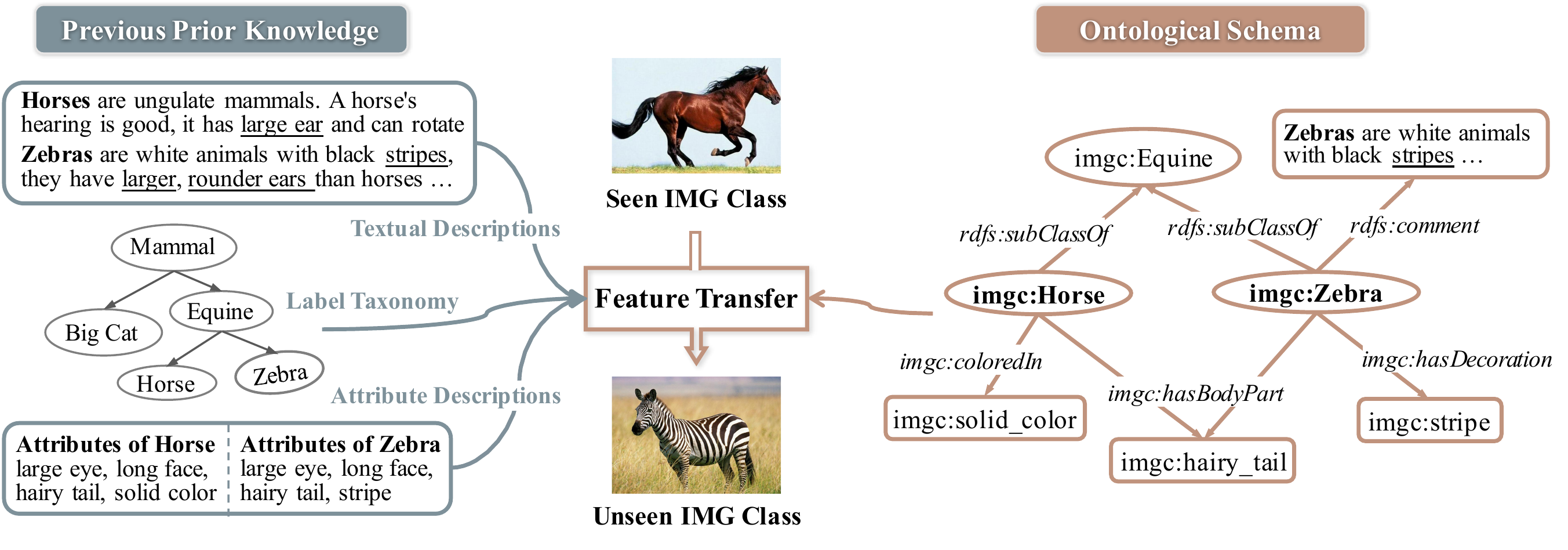}
\end{minipage}%
}%

\subfigure[Zero-shot Knowledge Graph Completion with Newly-added Relations]{
\begin{minipage}[c]{0.8\linewidth}
\centering
\includegraphics[width=14cm]{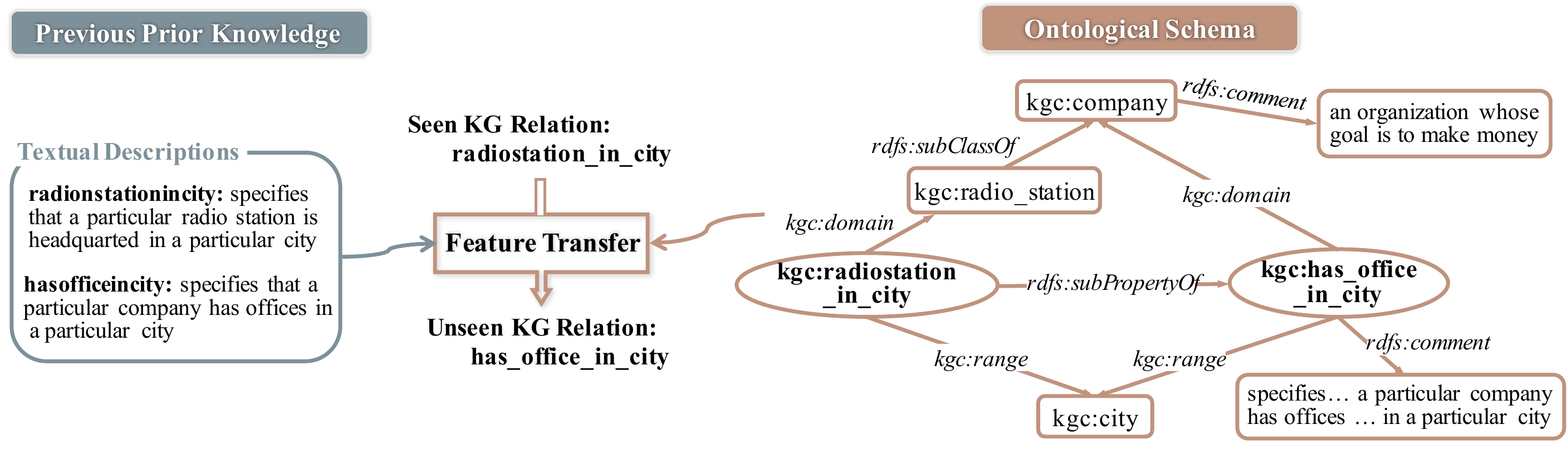}
\end{minipage}%
}%
\centering
\caption{Comparison of previously used prior knowledge [Left] and our proposed ontological schemas [Right] in zero-shot image classification and zero-shot knowledge graph completion tasks.
}\label{fig:example}
\end{figure*}

Inspired by the humans' abilities of recognizing new concepts only from their semantic descriptions and previous recognition experience, 
ZSL aims to develop models trained on data of seen classes and class semantic descriptions to make predictions on unseen classes.
These descriptions, also known as \textit{prior knowledge}, provide a prior about the semantic relationships between seen and unseen classes so that the model parameters (e.g., features) learned from seen classes can be transferred to unseen classes.
The majority of ZSL methods \cite{DBLP:conf/cvpr/ZhuEL0E18,frome2013devise,norouzi2013zero,Qin2020ZSGAN} consider textual descriptions as the priors.
For example, \cite{frome2013devise,norouzi2013zero} project images into the semantic embedding space pre-trained on textual corpora to classify unseen images.
\cite{Qin2020ZSGAN} generates relation embeddings for unseen relations from their text descriptions for embedding-based knowledge graph completion tasks.
Attribute descriptions \cite{lampert2013attribute,liu2020attribute,DBLP:conf/cvpr/LiJLD0H19} and label taxonomy \cite{DBLP:conf/cvpr/0004YG18,DBLP:conf/cvpr/KampffmeyerCLWZ19} are also widely adopted to model the semantic relationships between classes for zero-shot image classification.

\textbf{However, all of these priors are limited with incomplete semantics.}
As shown in the left of Figure~\ref{fig:example}, in general, textual descriptions may not provide distinguishing characteristics necessary for ZSL due to the noisy words in text \cite{DBLP:conf/cvpr/QiaoLSH16};
attribute descriptions focus on the ``local'' characteristics of objects and easily suffer from the domain shift problem \cite{DBLP:journals/pami/FuHXG15};
while label taxonomy merely considers the hierarchical relationships between classes.
 A more detailed discussion of these shortcomings is provided in Section~\ref{prior_know}.

We expect more expressive and competitive prior knowledge to boost the performance of ZSL.
In this study, we propose to utilize ontological schema which is convenient for defining the expressive semantics for a given domain  \cite{horrocks2008ontologies}.
An ontological schema models the general concepts (i.e., types of things) that exist in a domain and the properties that describe the semantic relationships between concepts.
It can also represent a variety of valuable information such as concept hierarchy and meta data (e.g., textual definitions, comments and descriptions of concepts), which facilitate modeling richer prior knowledge for ZSL.
The right of Figure~\ref{fig:example} shows some snapshots of such an ontological schema.

\textbf{In this paper, we propose a novel ZSL framework called OntoZSL which not only enhances the class semantics with an ontological schema, but also employs an ontology-based generative model to synthesize training samples for unseen classes.}
Specifically, an ontology embedding technique is first proposed to learn meaningful vector representations of ZSL classes (i.e., class embeddings) from the ontological schema. A generative model e.g., generative adversarial network (GAN) \cite{goodfellow2014generative} is then adopted to generate features and synthesize training samples for unseen classes conditioned on the class embeddings, empirically turning the ZSL problem into a standard supervised learning problem.

Many ontology embedding methods \cite{hao2019universal,chen2018on2vec,gutierrez2018knowledge} have been invented and extended from knowledge graph embedding (KGE) techniques \cite{wang2017knowledge}.
However, adapting existing KGE methods to encode ontologies for ZSL is problematic due to the inherent graph structure and lexical information that both exist in the ontological schema.
That is to say, each concept in the ontological schema may have two types of semantics, one is structure-based and captures the multi-relational structures, while the other is text-based and describes the concepts using some natural language tokens, e.g., the concept \textit{kgc:company} in Figure~\ref{fig:example}.
We thus propose a text-aware ontology embedding technique, which can learn the structural and textual representations for each concept simultaneously.

Developing generative models such as GANs for ZSL has been a popular strategy recently, and proven to be more effective and easier to generalize compared with traditional mapping-based ZSL methods (more comparisons are introduced in Section~\ref{zsl_methods}).
However, most of existing generative ZSL methods are merely built upon one type of priors such as textual or attribute descriptions.
While in our paper, we propose an ontology-based GAN to incorporate richer priors within ontological schemas to generate more discriminative sample features for ZSL.

We demonstrate the effectiveness of our framework on ZSL tasks from two different domains including a task of image classification (IMGC) from vision and a knowledge graph completion (KGC) task.
In IMGC, we build an ontological schema for image classes.
While in KGC, we generalize the unseen concepts to unseen relations and build an ontological schema for KG relations.
The examples of ontological schemas in two domains are shown in Figure~\ref{fig:example}.

Our main contributions are summarized as below:
\begin{itemize}

\item To the best of our knowledge, this is among the first ones to explore expressive class semantics from ontological schemas in Zero-shot Learning.

\item We propose OntoZSL, a novel ontology-guided ZSL framework that not only adopts a text-aware ontology embedding method to incorporate prior knowledge from ontological schemas, but also employs an ontology-based generative model to synthesize training samples for unseen classes.

\item In comparison with the state-of-the-art baselines, our framework achieves promising scores on image classification task for standard AwA \cite{xian2019survey} and constructed ImNet-A and ImNet-O datasets, as well as on knowledge graph completion task for datasets from NELL and Wikidata\footnote{Our code and datasets are available at \url{https://github.com/genggengcss/OntoZSL}.}.
\end{itemize}

\section{Preliminaries}
We first begin by formally introducing the zero-shot learning in two tasks: image classification (IMGC) and knowledge graph completion (KGC), and their corresponding ontological schemas used.

\subsection{Zero-shot Image Classification}
Zero-shot learning in image classification is to recognize the new classes whose images are not seen during training.
Let $\mathcal{D}_{tr} = \{(x, y) | x \in \mathcal{X}_s, y \in \mathcal{Y}_s\}$ be the training set, where $x$ is the CNN features of a training image, $y$ denotes its class label in $\mathcal{Y}_s$ consisting of seen classes.
While the testing set is denoted as $\mathcal{D}_{te} = \{(x, y) | x \in \mathcal{X}_u, y \in \mathcal{Y}_u\}$,
where $\mathcal{Y}_u$, the set of unseen classes, has no overlap with $\mathcal{Y}_{s}$.
Suppose that we have class representations $O \in \mathbb{R}^{n \times (|\mathcal{Y}_s|+|\mathcal{Y}_u|)}$ learned from semantic descriptions for $|\mathcal{Y}_s|$ seen classes and $|\mathcal{Y}_u|$ unseen classes, the task of zero-shot IMGC is to learn a classifier for each unseen class given $\{\mathcal{D}_{tr}, O\}$ for training.
These representations can be provided as binary/numerical attribute vectors, word embeddings/RNN features or class embeddings learned from our ontological schema.
We study two settings at the testing stage: one is standard ZSL which classifies the testing samples in $\mathcal{X}_u$ with candidates from $\mathcal{Y}_u$, while the other is generalized ZSL (GZSL) which extends the testing set to $\mathcal{X}_s \cup \mathcal{X}_u$, with candidates from both seen and unseen classes i.e., $\mathcal{Y}_s \cup \mathcal{Y}_u$.

\subsection{Zero-shot Knowledge Graph Completion}\label{kgc_formulation}

Different from the clustered instances in IMGC, a KG $\mathcal{G} =\{ \mathcal{E}, \mathcal{R}, \mathcal{T}\}$ is composed of a set of entities $\mathcal{E}$, a set of relations $\mathcal{R}$ and a set of triple facts $\mathcal{T} = \{(h, r, t)| h, t \in \mathcal{E}; r \in \mathcal{R}\}$.
The task of knowledge graph completion (KGC) is proposed to improve Knowledge Graphs by completing the triple facts in KGs when one of $h,r,t$ is missing.
Typical KGC methods utilize KG embedding models such as TransE \cite{bordes2013translating} to embed entities and relations in continuous vector spaces (e.g., the embeddings of $h,r,t$ are represented as $x_h, x_r, x_t$ respectively) and conduct vector computations to complete the missing triples, which are trained by existing triples and assume all testing entities and relations are available at training time.
Therefore, the zero-shot KGC task is defined as predicting for the newly-added entities or relations which have no associated triples in the training data.

In this study, we focus on those newly-added KG relations (i.e., unseen relations).
Specifically, we separate two disjoint relation sets: the seen relation set $\mathcal{R}_s$ and the unseen relation set $\mathcal{R}_u$.
The triple set $\mathcal{T}_s = \{(h, r_s, t)| h, t \in \mathcal{E}; r_s \in \mathcal{R}_s\}$ is then collected for training, and $\mathcal{T}_u = \{(h, r_u, t)| h, t \in \mathcal{E}; r_u \in \mathcal{R}_u\}$ is collected to evaluate the prediction of the triples of unseen relations.
It is noted that we consider a closed set of entities, i.e., each entity that appears in the testing set already exists in the training set, because making both entity set and relation set open makes the problem much more challenging, and the current work now only considers one of them (see references introduced in Section~\ref{zsl_kgc}).
% (e.g., open relations in \cite{Qin2020ZSGAN} and open entities in \cite{ShiW2018Open,zhao2020attention}).
Similar to IMGC, there are also semantic representations of relations in $R_s \cup R_u$, which are learned from textual descriptions or ontological schemas.

With the zero-shot setting, the KGC problem in our study is formulated as predicting the tail entity $t$ given the head entity $h$ and the relation $r$ in a triple.
More specifically, for each query tuple ($h, r$), we assume there is one ground-truth tail entity $t$ such that the triple $(h, r, t)$ is true\footnote{Generally in KGs, there may be more than one correct tail entity for a query tuple. Here, we follow previous KGC work \cite{wang2014knowledge} to apply a \textit{filter} setting during testing where other correct tail entities are filtered before ranking and only the current test one left.}.
The target of KGC model is to assign the highest ranking score to $t$ against the rest of all the candidate entities which are denoted as $ C_{(h, r)}$.
Therefore, during zero-shot testing, we will predict the triple facts of $r_u$ by ranking $t$ with the candidate tail entities $t' \in C_{(h, r_u)}$.
Accordingly, we do not set generalized ZSL (GZSL) testing in this case, considering that the candidate space only involves entities and the prediction with unseen relations is independent of the prediction with seen relations, while the latter is a traditional KGC task which is out of the scope of this paper.

\subsection{Ontological Schema}
Ontological schemas are used as the semantic prior knowledge for the above ZSL tasks in our paper.
The ontology, denoted as $\mathcal{O}=\{\mathcal{C}^O, \mathcal{P}^O, \mathcal{T}^O\}$, is a multi-relational graph formed with $\mathcal{C}^O$, a set of concept nodes, $\mathcal{P}^O$, a set of property edges, and $\mathcal{T}^O = \{(c_i, p, c_j)|c_i, c_j \in \mathcal{C}^O, p \in \mathcal{P}^O\}$, a set of RDF triples.
The concept nodes here refer to the domain-specific concepts.
For example, in IMGC, they are image classes, image attributes, etc. While in KGC, they are KG relations, and their domain (i.e., head entity types) and range (i.e., tail entity types) constraints, etc.
As for property edge, it refers to a link between two concepts.
The properties in our ontology are a combination of domain-specific properties (e.g., \textit{imgc:hasDecoration}) and RDF/RDFS\footnote{\url{https://www.w3.org/TR/rdf-schema/}} built-in properties (e.g., \textit{rdfs:subClassOf}, \textit{rdfs:subPropertyOf}).
For example, the triple (\textit{imgc:Zebra, imgc:hasDecoration, imgc:Stripe}) in Figure~\ref{fig:example} denotes that animal class ``Zebra'' is decorated with ``Stripe'',
while the triple (\textit{kgc:radiostation\_in\_city, rdfs:subPropertyOf, kgc:has\_office\_in\_city}) denotes that KG relation ``radiostation in city'' is a subrelation of ``has office in city''.  
In addition to RDF triples with structural relationships between concepts, each concept in the ontological schema also contains a paragraph of textual descriptions.
 These descriptions are lexically meaningful information of concepts, which can also be represented by triples using properties e.g., \textit{rdfs:comment}.
 
In our study, we use a semantic embedding technique to encode all the concept nodes in the ontological schema as vectors, through which the class labels in IMGC and the relation labels in KGC are embedded.
They are then used as the additional input of GAN models to generate more discriminative samples for unseen image classes or unseen KG relations.

\section{Related Work}

\subsection{Prior Knowledge for ZSL}\label{prior_know}

The first we discuss is the prior knowledge previously used in the ZSL literature.
Some works employ attribute descriptions as the priors \cite{DBLP:conf/cvpr/FarhadiEHF09,lampert2013attribute,DBLP:conf/cvpr/AkataPHS13,DBLP:conf/cvpr/LiJLD0H19,liu2020attribute}.
In these works, each class is annotated with a series of attributes which describe its characteristics, and the semantic relationships between classes are thus represented by those shared attributes.
However, attributes focus on describing ``local'' characteristics and the semantically identical attributes may perform inconsistently across different classes.
For example, in image classification, the animal classes ``Horse'' and ``Pig'' share the same attribute ``hasTail", but the visual appearance of their tails differs greatly.
The model trained with ``Horse'' may not generalize well on the prediction of ``Pig'' (i.e., domain shift problem mentioned earlier).
Some works prefer to utilize textual descriptions or distributed word embeddings of classes pre-trained on textual data to model the class semantics \cite{frome2013devise,norouzi2013zero,DBLP:conf/cvpr/ZhangXG17,Qin2020ZSGAN}.
Textual data can be easily obtained from linguistic sources such as Wikipedia articles, however, they are noisy and often lead to poor performance.

There are also some works utilizing label ontologies for inter-class relationships, such as label taxonomy \cite{DBLP:conf/cvpr/0004YG18,DBLP:conf/cvpr/KampffmeyerCLWZ19}, Hierarchy and Exclusion (HEX) label graph \cite{DBLP:conf/eccv/DengDJFMBLNA14}, and label ontology in OWL (Web Ontology Language) \cite{chen2020ontology}.
However, these ontologies also have their limitations.
The label taxonomy lacks discriminative semantics for those sibling classes which may look quite different (e.g., ``Horse'' and ``Zebra'' in Figure~\ref{fig:example}), while the HEX label graph still focuses on modeling the relationships between any two labels via attributes -- one class is regarded as a subclass of the attributes annotated for it, and as exclusive with those that are irrelevant with it.
Different from these works, our proposed ontological schema contains more complete semantics, in which the existing priors are well fused and benefit each other. For example, the class-level priors such as label taxonomy provide global constraint for attribute descriptions while class-specific attributes provide more detailed and discriminative priors for classes especially for sibling classes.

\textbf{Comparison with OWL-based label ontology \cite{chen2020ontology}.}
Although it expresses the same complete class semantics as we do, the OWL-based semantic representation is difficult to apply due to its complicated definition.
While our ontological schema is mainly in the form of multi-relational graphs composed of RDF triples, which is easier to model and embed using many successful triple embedding algorithms.
On the other hand, the construction of the ontologies used in \cite{chen2020ontology} heavily relies on the manual work, while our ontological schemas are built upon existing resources or are directly available.

\subsection{Zero-shot Learning Strategy}\label{zsl_methods}

Given the prior knowledge, existing approaches differ significantly in how the features are transferred from seen classes to unseen classes.
One branch is based on mapping. 
Some methods \cite{frome2013devise,lampert2013attribute,norouzi2013zero,DBLP:conf/cvpr/KodirovXG17} learn an instance-class mapping with seen samples in training.
In testing, the features of an input are projected into the vector space of the labels, and the nearest neighbor (a class label) in that space is computed as the output label.
Some other methods \cite{DBLP:conf/cvpr/ChangpinyoCGS16,DBLP:conf/cvpr/ZhangXG17,DBLP:conf/cvpr/0004YG18,DBLP:conf/cvpr/KampffmeyerCLWZ19} learn a reverse mapping -- labels are mapped to the space of input instances.
However, all of these mappings are trained by seen samples, and thus have a strong bias towards seen classes during prediction, especially in generalized ZSL where the output space includes both seen and unseen classes.

Recently, by taking advantages of generative models such as GANs, several methods  \cite{DBLP:conf/cvpr/XianLSA18,DBLP:conf/cvpr/ZhuEL0E18,DBLP:conf/cvpr/HuangWYW19,DBLP:conf/cvpr/LiJLD0H19,zhu2019learning,geng2020generative,Qin2020ZSGAN} have been proposed to directly synthesize samples (or features) for unseen classes from their prior knowledge, which convert the ZSL problem to a standard supervised learning problem with the aforementioned bias issue avoided.
Although these generative models are trained using the samples of seen classes, the generators can generalize well on unseen classes according to the semantic relationships between them.
In this study, we also introduce and evaluate GANs in our framework.
As far as we know, our work is among the first to incorporate the ontological schema with GAN for feature generation.

\subsection{Zero-shot Knowledge Graph Completion}\label{zsl_kgc}
Reviewing the literature of ZSL, we find that most of works especially those mentioned above are developed in the computer vision community for image classification.
There are also several ZSL studies for knowledge graph completion.
Some of them devote to deal with the unseen entities by exploiting the auxiliary connections with seen entities \cite{HamaguchiOSM2017OOKB,ShiW2018Open,wang2019logic,zhao2020attention}, introducing their textual descriptions \cite{shah2019open}, or learning entity-independent relational semantics which summarize the substructure underlying KG so that naturally generalizing to unseen entities \cite{teru2019inductive}.
While few works such as \cite{Qin2020ZSGAN} focus on the unseen relations.
In our work, we also concentrate on these newly-added relations.
Different from \cite{Qin2020ZSGAN} which generates unseen relation embeddings solely from their textual descriptions, our OntoZSL generates from the ontological schema which describes richer correlations between KG relations, such as domain and range constraints.
Besides, OntoZSL is well-suited for zero-shot KGC considering that many KGs inherently have ontologies which highly summarize the entities and relations in KGs.

\begin{figure}
\centering
\includegraphics[height=3.8cm]{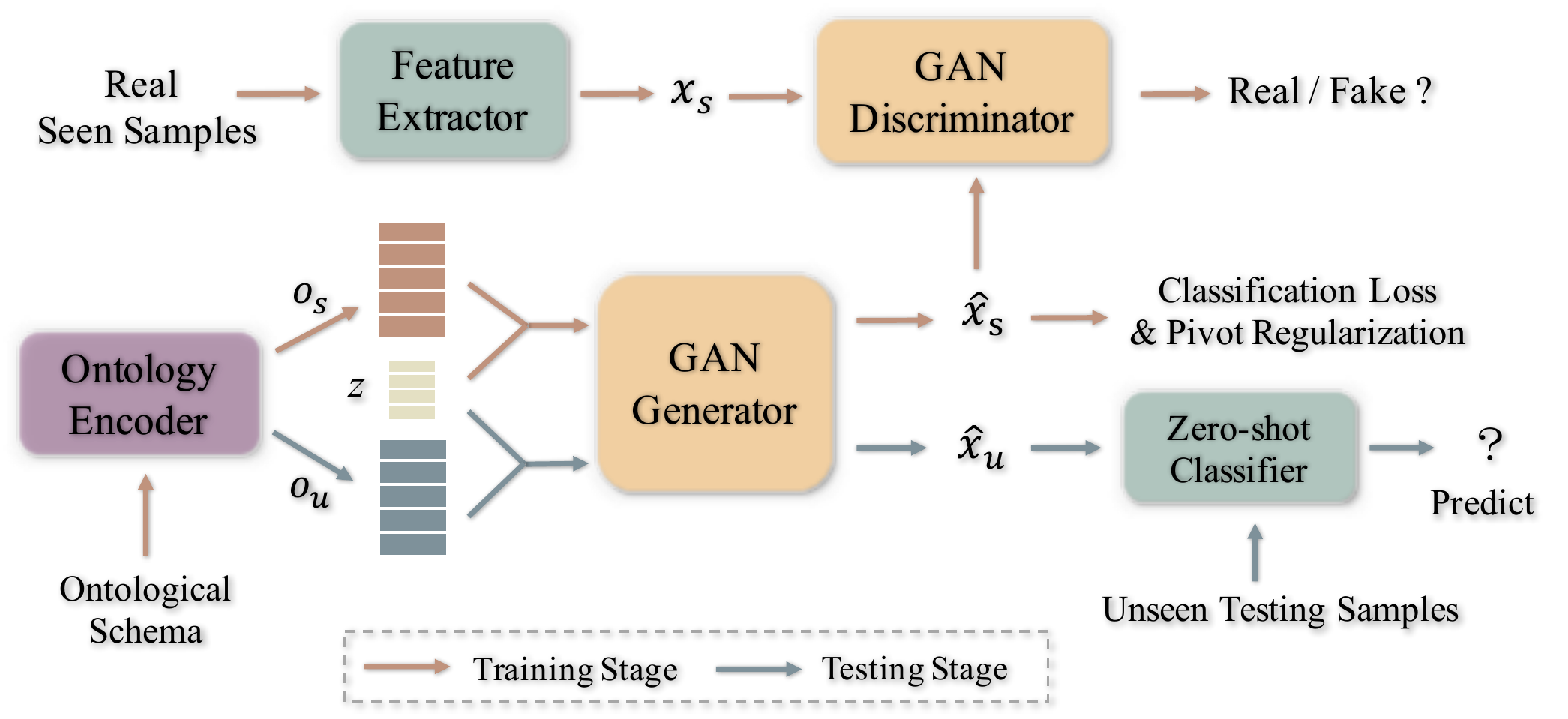}
\caption{An overview of OntoZSL in the standard ZSL setting. 
    $o_s$ and $o_u$ represent the semantic embeddings of seen and unseen concepts (i.e., image classes in IMGC task and KG relations in KGC task) learned from the ontological schema, respectively. During training, the GAN model generates fake samples $\hat{x}_s$ for seen concepts and distinguishes them with the real samples $x_s$ learned from a feature extractor.
    With a trained generator, the samples of unseen classes ($\hat{x}_u$) can be generated to learn classifiers for prediction.
    }
\label{fig:framework}
\end{figure}

\section{Methodology}
In this section, we will introduce our proposed general ZSL framework \textbf{OntoZSL}, which builds upon an ontology embedding technique and a generative adversarial network (GAN) and can be applied to two different zero-shot learning tasks: image classification (IMGC) and knowledge graph completion (KGC).
Figure~\ref{fig:framework} presents its overall architecture, including four core parts:

\textbf{Ontology Encoder.}
We learn semantically meaningful class representations or relation representations from the ontological schema using an ontology embedding technique, which considers the structural relationships between concepts as well as their correlations implied in textual descriptions.

\textbf{Feature Extractor.}
We utilize a feature extractor to extract the real data (instance) representations, which will be taken as the guidance of adversarial training.
Regarding the different data forms in two tasks, we adopt different strategies to obtain real data distributions for different tasks.

\textbf{Generation Model.}
A typical scheme of GAN is adopted for data generation.
It consists of (i) a generator to synthesize instance features from random noises,
(ii) a discriminator to distinguish the generated features from real ones,
and (iii) some additional losses to ensure the inter-class discrimination of generated features. 
Notably, we generate instance features instead of raw instances for both higher accuracy and computation efficiency \cite{DBLP:conf/cvpr/XianLSA18}.

\textbf{Zero-shot Classifier.}
With the well-trained generator, the samples of unseen classes (relations) can be synthesized with their semantic representations, from which the unseen classifiers can be learned to predict the testing samples of unseen classes (relations).

Among these parts, the ontology encoder and the generation model are general across different tasks, while the feature extractor and the zero-shot classifier is task-specific.
% In the following of this section,
Next, 
we will first introduce these parts w.r.t. the IMGC task, and then introduce the difference of the task-specific parts in addressing the KGC task.

\subsection{Ontology Encoder}\label{class_embed}
% A text-aware semantic embedding technique is proposed to encode the graph structure as well as textual information in the ontological schema.
In this subsection, we provide a text-aware semantic embedding technique for encoding the graph structure as well as textual information in the ontological schema.

\textbf{Default Embedding.}
Considering the structural RDF triples in ontological schema, there are many triple embedding techniques that can be applied \cite{wang2017knowledge}.
Given a triple $(c_i, p, c_j)$, the aim of triple embedding is to design a scoring function $f(c_i, p, c_j)$ as the optimization objective.
A higher score indicates a more plausible triple.
In this paper, we adopt a mature and widely-used triple embedding method TransE \cite{bordes2013translating} which assumes the property in each triple as a translation between two concepts.
Its score function is defined as follows:
\begin{equation}
	f_{TransE}(c_i, p, c_j) = -|| \bm{c}_i + \bm{p} - \bm{c}_j||
\end{equation}
where $\bm{c}_i, \bm{p}, \bm{c}_j$ denote the embeddings of $c_i, p, c_j$, respectively.
% $\circ$ is the Hadamard product and $\cdot$ is the dot product.

To learn the embeddings of all concepts in the ontological schema $\mathcal{O}$, a hinge loss is minimized for all triples in ontology:
\begin{equation}
\mathcal{J}_\mathcal{O} = \frac{1}{|\mathcal{T}^O|} \sum_{(c_i, p, c_j) \in \mathcal{T}^O \atop \wedge (c'_i, p, c'_j) \notin \mathcal{T}^O} [\gamma_o + f(c'_i, p, c'_j) - f(c_i, p, c_j)]_+
\end{equation}
where $\gamma_o$ is a margin parameter which controls the score difference between positive and negative triples, the negative triples are generated by replacing either head or tail concepts in positive triples with other concepts and not exist in the ontology.
Notably, there are other triple embedding techniques can potentially be used for encoding our ontological schema. Since exploring different techniques is not the focus of this paper, we leave them as future work.

\begin{figure}
\centering
\includegraphics[height=4.6cm]{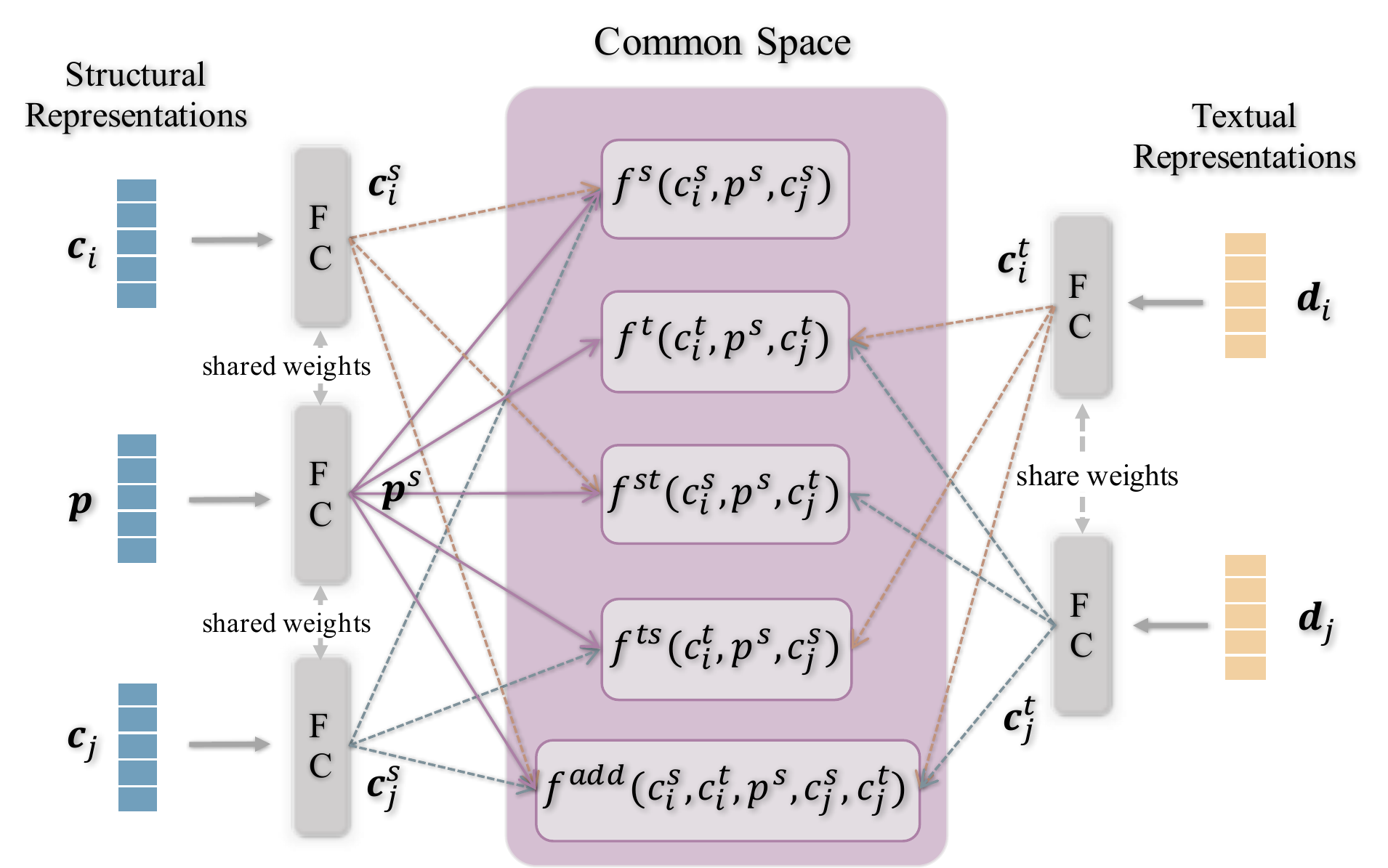}
\caption{Overview of the network architecture for text-aware semantic embedding.}
\label{fig:ontoE}
\end{figure}

\textbf{Text-Aware Embedding.}
However, the textual descriptions of concepts in ontological schema describe the knowledge of concepts from another modal.
Such semantics require special modeling than regular structural triples.
Therefore, we propose the text-aware semantic embedding model by projecting the structural representations and the textual representations into a common space and learning them simultaneously using the same objective score function, as shown in Figure~\ref{fig:ontoE}.

Specifically, given a triple $(c_i, p, c_j)$, we first project its structural embeddings $\bm{c}_i, \bm{p}, \bm{c}_j$ learned above and the textual representation of concepts $\bm{d}_i, \bm{d}_j$ into a common space using fully connected (FC) layers, e.g., $\bm{c}_i$ into $\bm{c}_i^s$ and $\bm{d}_i$ into $\bm{c}_i^t$ (cf. Figure~\ref{fig:ontoE}).
In this space, the structure-based score is still defined as proposed by TransE:
\begin{equation}
f^s = -|| \bm{c}_i^s + \bm{p}^s - \bm{c}_j^s ||
\end{equation}
while the text-based score is defined as:
\begin{equation}
f^t = -|| \bm{c}_i^t + \bm{p}^s - \bm{c}_j^t ||
\end{equation}
which also constrains the textual representations under the translational assumption.

To make these two types of representations compatible and complementary with each other, we follow the proposals of \cite{xie2016representation,xie2017image,mousselly2018multimodal} to define the crossed and additive score function:
\begin{equation}
\begin{split}
	f^{st} &= -|| \bm{c}_i^s + \bm{p}^s - \bm{c}_j^t ||
	\\
	f^{ts} &= -|| \bm{c}_i^t + \bm{p}^s - \bm{c}_j^s ||
	\\
	f^{add} &= -|| (\bm{c}_i^s + \bm{c}_i^t) + \bm{p}^s - (\bm{c}_j^s + \bm{c}_j^t) ||
\end{split}
\end{equation}
All of these score functions ensure that the two kinds of concept representations are learned in the same vector space.
The overall score function are defined as:
\begin{equation}
f^{T}(c_i, p, c_j) = 
	f^s + f^t + f^{st} + f^{ts} + f^{add}
\end{equation}

Therefore, the final training loss changes to:
\begin{equation}
\mathcal{J}_\mathcal{O}^{Text} = \frac{1}{|\mathcal{T}^\ast|} \sum_{(c_i, p, c_j) \in \mathcal{T}^{\ast} \atop \wedge (c'_i, p, c'_j) \notin \mathcal{T}^\ast} [\gamma_o + f^{T}(c'_i, p, c'_j) - f^{T}(c_i, p, c_j)]_+
\end{equation}
where $\mathcal{T}^\ast$ refers to the triple set with regular structural properties.
The triples with \textit{rdfs:comment} property that connects a concept with its textual description are removed here considering that these text nodes are encoded from another view.

After training, for each concept $i$ in ontological schema, we can learn two types of concept embeddings, i.e., structure based $\bm{c}_i^s$ and text based $\bm{c}_i^t$. To fuse the semantic features from these two types, we concatenate them to form the final concept embedding:
\begin{equation}
	o_i = [\bm{c}_i^s; \bm{c}_i^t] 
\end{equation}

As for the initial textual representations, we use word embeddings pre-trained on textual corpora to represent the words in the text.
Also, to suppress the text noises, we employ TF-IDF features \cite{salton1988term} to evaluate the importance of each word. Besides, the FC layers used for projection also have a promotion for noise suppression.

\subsection{Feature Extractor}
Following most of the previous works \cite{xian2019survey}, we employ ResNet101 \cite{HeZRS2016resnet} to extract the real features of images in zero-shot image classification.
ResNet is a well-performed CNN architecture pre-trained on ImageNet~\cite{DBLP:conf/cvpr/DengDSLL009} with 1K classes, in which no unseen classes of our evaluation datasets are involved.

\subsection{Feature Generation with GAN}
With class embeddings learned from ontology encoder, we train a generator $G$, which takes a class embedding $o_i$ and a random noise vector $z$ sampled from Normal distribution $\mathcal{N}(0,1)$ as input, and generates the CNN features $\hat{x}$ of class $i$.
The loss of $G$ is defined as:
\begin{equation}
	\mathcal{L}_G = - \mathbb{E}[D(\hat{x})] + \lambda_1 \mathcal{L}_{cls} (\hat{x}) + \lambda_2 \mathcal{L}_{P}
\end{equation}
where $\hat{x} = G(z, o_i)$. 
The first term of loss function is the Wasserstein loss \cite{arjovsky2017wasserstein} which can effectively eliminate the mode collapse problem during generation.
While the second term is a supervised classification loss for classifying the synthesized features, and the third item is a pivot regularization proposed by \cite{DBLP:conf/cvpr/ZhuEL0E18}, which regularizes the mean of generated features of each class to be the mean of real feature distribution.
Both of the latter two loss terms encourage the generated features to have more inter-class discrimination.
$\lambda_1$ and $\lambda_2$ are the corresponding weight coefficients.

The discriminator $D$ then takes the synthesized features $\hat{x}$ and the real features $x$ extracted from a training image of class $i$ as input, the loss can be formulated as:
\begin{equation}
\begin{aligned}
	\mathcal{L}_D = \mathbb{E}[D(x, o_i)] - \mathbb{E}[D(\hat{x})] \\- \beta \mathbb{E}[(|| \bigtriangledown_{\tilde{x}} D(\tilde{x}) ||_p -1)^2]
	\end{aligned}
\end{equation}
where the first two terms approximate the Wasserstein distance of the distribution of real features and synthesized features, and the last term is the gradient penalty to enforce the gradient of $D$ to have unit norm (i.e., Lipschitz constraint proposed by \cite{DBLP:conf/nips/GulrajaniAADC17}), in which $\tilde{x} = \varepsilon x + (1-\varepsilon) \hat{x}$ with $\varepsilon \sim U(0,1)$, and $\beta$ is the weight coefficient.

The GAN is optimized by a minimax game, which minimizes the loss of $G$ but maximizes the loss of $D$.
We also note that the generator and discriminator are both incorporated with the class embeddings during training.
This is a typical method of conditional GANs \cite{mirza2014conditional} that introduces external information to guide the training of GANs, which is consistent with the generative ZSL -- synthesizing instance features based on the prior knowledge of classes.

\subsection{Zero-shot Classifiers}
Once the GAN is trained to be able to generate sample features for seen classes, it can also synthesize features for unseen classes with random noises and their corresponding class embeddings.
Consequently, with synthesized unseen data $\hat{\mathcal{X}_u}$, we can learn a typical softmax classifier for each unseen class  and classify its testing samples.
The classifier is optimized by:
\begin{equation}
	\min \limits_\theta - \frac{1}{|\mathcal{X}|}  \sum_{(x,y) \in (\mathcal{X}, \mathcal{Y})} logP(y|x; \theta)
\end{equation}
where $\mathcal{X}$ represents the features for training, $\mathcal{Y}$ is the label set to be predicted on, $\theta$ is the training parameter and $P(y|x;\theta) = \frac{exp(\theta_y^Tx)}{\sum_i^{|\mathcal{Y}|} exp(\theta_i^Tx)}$.
Regarding the different prediction setting in IMGC, $\mathcal{X}=\hat{\mathcal{X}_u}$ when it is standard ZSL and $\mathcal{X}=\mathcal{X}_s \cup \hat{\mathcal{X}_u}$ when it is GZSL, while the label set $\mathcal{Y}$ corresponds to $\mathcal{Y}_u$ and $\mathcal{Y}_s \cup \mathcal{Y}_u$ respectively.

\begin{table*}[]
\small
\caption{Statistics of the zero-shot image classification datasets.
}
  \label{tab:datasets_imgc}
\begin{tabular}{c|c|ccc|ccccc|ccc}
\hline
% \toprule[0.6pt]
\multicolumn{1}{c|}{\multirow{3}{*}{\bf Datasets}} &
\multicolumn{1}{c|}{\multirow{3}{*}{\bf Granularity}}
&
\multicolumn{3}{c|}{\multirow{2}{*}{\bf \# Classes}} & \multicolumn{5}{c|}{\bf \# Images} & \multicolumn{3}{c}{\multirow{2}{*}{\bf Ontological Schema}}
\\
\multicolumn{1}{c|}{} 
& \multicolumn{1}{c|}{} 
& \multicolumn{3}{c|}{}           & 
& \multicolumn{2}{c}{\bf for Training} & \multicolumn{2}{c|}{\bf for Testing} & \multicolumn{3}{c}{}                                    \\ 
\multicolumn{1}{c|}{}
& \multicolumn{1}{c|}{}
& \multicolumn{1}{l}{\bf Total}   
& Seen  & Unseen  
&{\bf Total} & Seen & Unseen   & Seen   & Unseen
& \multicolumn{1}{l}{\# RDF Triples} & \# Concepts & \# Properties
\\
 \hline
AwA & coarse & 50 & 40 & 10 
& 37,322  & 23,527 & 0 & 5,882 & 7,913
& 1,256   & 180    &  12
\\
ImNet-A & fine   & 80 
% & 25 & 55 
& 28 & 52
% & 77,173  & 32,500 & 0 & 1,250  & 43,423
& 77,323 & 36,400 & 0 & 1,400 & 39,523
& 563 & 227 & 19    
\\
ImNet-O & fine  & 35 & 10 & 25 
& 39,361  & 12, 907 & 0 & 500  &  25,954
& 222 & 115 & 8         
\\
\hline
% \bottomrule[0.6pt]
\end{tabular}
\end{table*}

\subsection{Adapting to Knowledge Graph Completion}
Similar to zero-shot image classification, we can also generate features for unseen relations in knowledge graph with their semantic representations learned from ontological schema using the above generation model.
However, considering the different data instances in KGs, we adopt different strategies to extract sample features and design different zero-shot classifiers.

\textbf{Feature Extractor.}
Unlike traditional KG embedding methods which learn entity and relation embeddings based on some assumptions (or constraints), we hope to learn cluster-structure feature distribution for seen and unseen relation facts so that preserving the higher intra-class similarity and relatively lower inter-class similarity as most ZSL works did.
Therefore, we follow \cite{Qin2020ZSGAN} to learn and train the real relation embeddings in bags.
To be more specific, suppose that there are bags $\{B_r | r \in \mathcal{R}_s \}$ in the training set, a bag $B_r$ is named by a seen relation $r$, in which all the triples involving relation $r$ are contained.
The real embeddings $x_r$ of relation $r$ are thus represented by the embeddings of entity pairs in bag $B_r$, whose training are thus supervised by some reference triples in this bag.

Concretely, for an entity pair ($h, t$) in bag $B_r$, we first embed each entity using a simple fully connected (FC) layer and generate the entity pair embedding $u_{ep}$ as:
\begin{equation}
\begin{split}
u_{ep} &= \sigma([f_1(x_h);f_1(x_t)])
\\
f_1(x_h) &= W_1(x_h) + b_1
\\
f_1(x_t) &= W_1(x_t) + b_1
\end{split}
\end{equation}
where $[\cdot ; \cdot]$ represents the vector concatenation operation, $\sigma$ is the \textit{tanh} activation function.
We also consider the one-hop structure of each entity.
For the tail entity $t$, its structural embedding $u_t$ is represented as:
\begin{equation}
\begin{split}
	u_t = \sigma(\frac{1}{|\mathcal{N}_t|}
	\sum_{(r^n, t^n) \in \mathcal{N}_t} f_2(x_{r^n}, x_{t^n})),
	\\
	f_2(x_{r^n}, x_{t^n}) = W_2([x_{r^n}; x_{t^n}]) + b_2
\end{split}
\end{equation}
where $\mathcal{N}_t = \{(r^n, t^n)|(t, r^n, t^n) \in \mathcal{T}_s\}$ denotes the one-hop neighbors of entity $t$, and $f_2$ is a FC layer which encodes the neighborhood information.
In consideration of the scalability, the number of neighbors (i.e., $\left| \mathcal{N}_t \right|$) is set with an upper limit e.g., $50$.
The structural embedding of the head entity $h$, denoted as $u_h$ is calculated in the same way as the tail entity.
The final entity pair embedding (i.e., the relation embedding $x_r$) is then formulated as:
\begin{equation}
x_r = x_{(h,t)} = [u_{ep}; u_h; u_t]
\end{equation}

We train the real relation embeddings with some reference triples.
Specifically, for each relation $r$, the triples in bag $B_r$ are randomly split into two parts: one is taken as the reference set $\{h^{\ast}, r, t^{\ast}\}$, and the other is taken as the positive set $\{h^{+}, r, t^{+}\}$.
We also generate a set of negative triples $\{h^{+}, r, t^{-}\}$ by replacing the tail entity of each triple in the positive set with other entities.
With $m$ reference triples, we take the mean of reference relation embeddings, i.e., $x_{(h^{\ast}, t^{\ast})}^c = \frac{1}{m} \sum_{i=1}^{m} x_{(h^{\ast}, t^{\ast})}^i$, where $x_{(h^{\ast}, t^{\ast})}^i$ is computed by equations (12), (13) and (14), and calculate its cosine similarity with the relation embedding of each positive triple (i.e., $x_{(h^{+}, t^{+})}$) as a positive score denoted as $score^+$,
and calculate its cosine similarity with that of each negative triple (i.e., $x_{(h^{+}, t^{-})}$) as a negative score denoted as $score^-$.
A hinge loss is then adopted to optimize the training:
\begin{equation}
\mathcal{J}_{FE} = \frac{1}{|B_r^\ast|} \sum_{(h^{+}, r, t^{+}) \in B_r^\ast \atop \wedge (h^{+}, r, t^{-}) \notin B_r^\ast} [\gamma_f + score^+ - score^-]_+
\end{equation}
where $B_r^\ast$ means the training triples of relation $r$ except reference triples, and $\gamma_f$ denotes the margin parameter.
Instead of random initialization, we use pre-trained KG embedding to initialize the entities and relations in bags and neighborhoods.

During feature generation, when generating the fake relation embedding $\hat{x}_r = G(z, o_r)$ for relation $r$, we also take the above hinge loss as the classification loss to preserve the inter-class discrimination.
Specifically, a positive score is calculated between $\hat{x}_r$ and the cluster center of real relation embeddings, i.e., $x_r^c = \frac{1}{|N_r|} \sum_{i=1}^{N_r} x_r^i$, where $N_r$ denotes the number of training triples of relation $r$.
A negative score is computed between $x_r^c$ and the negative relation embedding calculated by negative triples with replaced tail entities.

\textbf{Zero-shot Classifiers.}
With the well-trained generator, we can generate plausible relation embedding $\hat{x}_{r_u} = G(z, o_{r_u})$ for unseen relation $r_u$ with its semantic representations $o_{r_u}$.
For a query tuple ($h, r_u$), the similarity ranking value $v_{(h, r_u, t')}$ of candidate tail $t'$ is calculated by the cosine similarity between $
\hat{x}_{r_u}$ and $x_{(h, t')}$.
The candidate with the highest value is the predicted tail entity of tuple ($h, r_u$).
For better generalization, we generate multiple relation embeddings for each relation and average the ranking value:
\begin{equation}
v_{(h, r_u, t')} = \frac{1}{|N_{g}|} \sum_{i=1}^{N_{g}} cosine(\hat{x}_{r_u}^i, x_{(h, t')})
\end{equation}
where $N_{g}$ denotes the number of generated relation embeddings for relation $r_u$.

\section{Experiments}
In the experiments, we evaluate OntoZSL by the two different tasks of zero-shot image classification and zero-shot knowledge graph completion.
We also compare the ontological schema against other prior knowledge for zero-shot learning, and finally analyze the impact of different semantics of the ontological schema.

\subsection{Image Classification}\label{sec:imgc_experiments}

\textbf{Datasets.}
We evaluate the zero-shot image classification task with a standard benchmark named Animals with Attributes (AwA) and two benchmarks ImNet-A and ImNet-O which are contributed by ourselves.
AwA \cite{xian2019survey} is a widely used coarse-grained benchmark for animal classification which contains $50$ animal classes with $37,322$ images, while ImNet-A and ImNet-O are two fine-grained datasets we extract from ImageNet \cite{DBLP:conf/cvpr/DengDSLL009}.
ImNet-A is for the classification of animals, while ImNet-O is for the classification of more general objects.
Details of the construction of the latter two benchmarks can be found in Appendix~\ref{datasets_attribute_annotation}.
The classes of all the three benchmarks are split into seen and unseen classes, following the {\it seen-unseen} strategy proposed in \cite{xian2019survey}.

Table~\ref{tab:datasets_imgc} provides detailed statistics of these datasets.
Compared with AwA, ImNet-A and ImNet-O are more challenging as they have fewer seen classes.
Both of these datasets have inherent label taxonomies which are helpful for building the ontological schemas.
Specifically, each class corresponds to a node in WordNet \cite{miller1995wordnet} -- a lexical database of semantic relations between words, and thus these classes are underpinned by the same taxonomy as WordNet.

\begin{table*}[]
\small
\caption{The $acc$uracy ($\%$) of image classification in the standard and generalized ZSL settings. The best results are marked in bold. ``--'' means the case where the method cannot be applied.
}
\label{tab:imgc_results}
\begin{tabular}{l|ccc|ccc|ccc|ccc}
\hline
% \toprule[0.6pt]
\multicolumn{1}{c|}{\multirow{3}{*}{\bf Methods}} 
& \multicolumn{3}{c|}{\bf Standard ZSL}               & \multicolumn{9}{c}{\bf Generalized ZSL}             \\
 & {\bf AwA}  & {\bf ImNet-A}  & {\bf ImNet-O}               
    & \multicolumn{3}{c|}{\bf AwA} 
    & \multicolumn{3}{c|}{\bf ImNet-A} 
    & \multicolumn{3}{c}{\bf ImNet-O}                       
    \\
& \multicolumn{1}{c}{$acc$} 
& \multicolumn{1}{c}{$acc$} 
& \multicolumn{1}{c|}{$acc$} 

& \multicolumn{1}{c}{$acc_s$}                      
& \multicolumn{1}{c}{$acc_u$} 
& \multicolumn{1}{c|}{$H$} 

& \multicolumn{1}{c}{$acc_s$}     
& \multicolumn{1}{c}{$acc_u$}   
& \multicolumn{1}{c|}{$H$} 
 
& \multicolumn{1}{c}{$acc_s$}     
& \multicolumn{1}{c}{$acc_u$}
& \multicolumn{1}{c}{$H$}    
\\
\hline
% \midrule[0.5pt]
\ DeViSE  
& 37.46  
% & 14.56 
& 14.30
& 14.32                 
& 81.06  & 3.29   & 6.32
% & 69.20 & 0.85 & 1.68
& 60.21 & 0.64 & 1.27
& 68.00 & 3.68 & 6.98 
\\
\ CONSE  
& 22.99  & 20.28  & 12.41                 
& 51.64  & 3.28   & 6.17                       
& 86.40  & 0.00  & 0.00                     
& 62.00  & 0.00  & 0.00                    
\\
\ SAE               
& 42.28 
% & 18.84
& 18.98
& 14.84                 
& 71.03 & 9.79   & 17.21                      
% & 86.48  & 0.00  & 0.00  
& 84.43 & 0.17 & 0.34
& {\bf 92.60}  & 0.16  & 0.32                     
\\
\ SYNC                     
& 39.14  & 20.52  & 18.58 
& {\bf 88.21}  & 8.43  & 15.39  
& {\bf 88.72} & 0.00 & 0.00 
& 62.53 & 0.00 & 0.00 
\\
\hline
% \midrule[0.5pt]
\ GCNZ                     
& -- 
% & 31.62
& 32.47
& 30.05                 
& -- & --   & --    
% & 49.52  & 15.68  & 23.81
& 47.79 & 15.15 & 23.01
& 44.60  & 14.48  & 21.87                    
\\
\ DGP                      
& 58.99  
% & 33.07
& 34.88
& 31.23                 
& 86.19  & 16.59  & 27.82                
% & 44.72  & 17.92  & 25.59
& 50.14 & 17.87 & 26.35
& 47.40  & 19.00  & 27.13                    
\\
\hline
% \midrule[0.5pt]
\ GAZSL                    
& 56.29  
%& 20.57
& 21.20
& 19.40                 
& 87.64  & 15.40  & 26.19                 
& 86.56  & 1.28   & 2.52                     
& 86.80  & 6.16   & 11.50                    
\\
\ LisGAN                   
& 58.89  
%& 21.00
& 21.90
& 20.20                 
& 60.03  & 44.30  & 50.98
% & 39.84  & 13.82  & 20.52
& 35.50 & 15.55 & 21.62
& 35.00  & 13.87  & 19.87                    
\\
\ LsrGAN  
& 56.34  
%& 19.56
& 19.69
& 20.20                
& 85.98  & 35.73  & 50.48  
% & 41.44  & 13.99  & 20.92
& 36.29 & 13.49 & 19.67
& 37.80  & 14.27  & 20.72                   
\\
\hline
% \midrule[0.5pt]
OntoZSL         
& {\bf 63.31}  
% & {\bf 38.72} 
& {\bf 39.00}
& {\bf 34.24}
& 64.90  & {\bf 49.35}  & {\bf 56.06}                
% & 37.92  & {\bf 26.65}  & {\bf 31.30}
& 37.86  & {\bf 27.94}  & {\bf 32.15}
& 43.40  & {\bf 21.50}  & {\bf 28.76}
% \\
% OntoZSL-Mult        
% &        &       &         
% &        &       &                     
% &        &       &                         
% &        &       &
\\
\hline                         
% \bottomrule[0.7pt]
\end{tabular}
\end{table*}

\textbf{The Ontological Schema for IMGC} mainly focuses on the class hierarchy, class attributes and literal descriptions.
To build such an ontological schema, we first adopt the taxonomy of WordNet to define the class hierarchy, where the class concepts are connected via the property {\it rdfs:subClassOf}, as Figure~\ref{fig:example} shows.
Then, we define the domain-specific properties such as {\it imgc:hasDecoration}, {\it imgc:coloredIn} to associate the class concepts with attribute concepts so that describing the visual characteristics of classes in ontology.
The attributes of classes are usually hand-labeled, in our paper, we reuse existing attribute annotations for AwA \cite{lampert2013attribute} and manually annotate attributes for classes in ImNet-A and ImNet-O since they have no open attributes.
During annotating, we also transfer some annotations from other datasets (e.g., AwA) to reduce the annotation cost (more details are in Appendix~\ref{datasets_attribute_annotation}).
As for the literal descriptions of concepts, we adopt the words of class names, which are widely-used text-based class semantics in the literature.
The statistics of constructed ontological schemas are shown in Table~\ref{tab:datasets_imgc}.

\textbf{Baselines and Metrics.}
We compare our framework with classic ZSL methods published in the past few years and the state-of-the-art ones reported very recently.
Specifically, {\bf DeViSE} \cite{frome2013devise}, {\bf CONSE} \cite{norouzi2013zero}, {\bf SAE} \cite{DBLP:conf/cvpr/KodirovXG17} and {\bf SYNC} \cite{DBLP:conf/cvpr/ChangpinyoCGS16} are mapping-based which map the image features into the label space represented by class embeddings or vice versa;
while {\bf GAZSL} \cite{DBLP:conf/cvpr/ZhuEL0E18}, {\bf LisGAN} \cite{DBLP:conf/cvpr/LiJLD0H19} and {\bf LsrGAN} \cite{VyasVPECCV20LsrGAN} are generative methods which generate visual features conditioned on the class embeddings.
We evaluate these methods with their available class embeddings.
For AwA, the binary attribute vectors are adopted, while for ImNet-A and ImNet-O, as the attributes of ImageNet classes are not available, we use the word embeddings of class names provided by \cite{DBLP:conf/cvpr/ChangpinyoCGS16}.
We also make a comparison with {\bf GCNZ} \cite{DBLP:conf/cvpr/0004YG18} and {\bf DGP} \cite{DBLP:conf/cvpr/KampffmeyerCLWZ19} which leverage the word embeddings of class labels and label taxonomy to predict on AwA and ImageNet.

% We measure the performance of the methods by accuracy.
We evaluate these methods by accuracy.
Considering the imbalanced samples across classes, we follow the current literature \cite{xian2019survey} to report the class-averaged (macro) accuracy.
Specifically, we first calculate the per-class accuracy -- the ratio of correct predictions over all the testing samples of this class, and then average the accuracies of all targeted classes as the final metric.
% \begin{equation}
% 	acc = \frac{1}{|\mathcal{Y}|} \sum_{c=1}^{|\mathcal{Y}|} \frac{\# \  \text{correct predictions in c}}{\# \  \text{images in c}}
% \end{equation}
Regarding the two testing settings in zero-shot image classification, the metrics are computed on all unseen classes in the standard setting, while in the GZSL setting, the class-averaged accuracy is calculated on seen and unseen classes separately, denoted as $acc_s$ and $acc_u$ respectively, and then a harmonic mean $H = (2 \times acc_s \times acc_u)/(acc_s + acc_u)$ is computed as the overall metric.

\textbf{Implementation.}
We employ ResNet101 \cite{HeZRS2016resnet} to extract $2,048$-dimensional visual features of images.
As for the word embeddings used for initializing the textual representation of ontology concepts, we use released $300$-dimensional word vectors, which are pre-trained on Wikipedia corpora using GloVe \cite{DBLP:conf/emnlp/PenningtonSM14} model.

The results are reported based on the following settings.
For ontology encoder, we set $\gamma_o=12$ as default, and learn the class embedding of dimension $200$ (i.e., $100$-dimensional structure-based representation and $100$-dimensional text-based representation).
We set other parameters in ontology encoder as recommended by \cite{SunDNT19Rotate} and \cite{mousselly2018multimodal}.
Regarding GAN, the generator and discriminator both consist of two fully connected layers.
The generator has $4,096$ hidden units and outputs image features with $2,048$ dimensions, while the discriminator also has $4,096$ hidden units and outputs a $2$-dimensional vector to indicate whether the input feature is real or not.
The dimension of noise vector $z$ is set to $100$. 
The learning rate is set to $0.0001$.
The weight $\lambda_1$ for classification loss is set to $0.01$, $\lambda_2$ for pivot regularization is set to $5$, and the weight $\beta$ for gradient penalty is set to $10$.

\textbf{Results.}
We report the prediction results under the standard ZSL setting and the generalized ZSL setting in Table~\ref{tab:imgc_results}.
Giving a first look at the standard ZSL, we find 
that our method achieves the best accuracy on all three datasets.
In comparison with the mapping-based baselines, e.g., DeVise, CONSE and the generative baselines, e.g., GAZSL, LsrGAN, our method outperforms the traditional class attribute annotations used for AwA as well as class word embeddings for ImNet-A and ImNet-O.
Most importantly, it also has a better performance than the previously proposed label ontologies (i.e., GCNZ and DGP), which simply consider the hierarchical relationships of classes.
These observations demonstrate the superiority of our OntoZSL compared with the state of the art.

While in the GZSL setting, we have similar observations as the standard one.
Our method performs better than the baselines and obtains significant outperformance on the metrics of $acc_u$ and $H$.
This shows our method has a better generalization.
Furthermore, we notice that among all the methods which utilize the same prior knowledge (i.e., word embeddings of classes or class attribute vectors), the performance of those mapping-based ones dramatically drops in comparison with the standard ZSL setting. CONSE and SYNC even drop to $0.00$ on ImNet-A and ImNet-O.
This verifies our points that these methods have a bias towards seen classes during prediction, i.e., their models tend to predict the unseen testing samples on seen classes.
In contrast, those generative methods which generate training samples for unseen classes have no such bias towards unseen classes.
We also find that although our framework does not achieve the best results on the prediction of seen testing samples ($acc_s$), it still accomplishes competitive performance as the state-of-the-arts.
This motivates us to explore algorithms to predict unseen testing samples correctly as well as maintain reasonable accuracy on seen classes.

% \todo{[TODO:]}
% Comparing the different ontology embedding score functions used in our method,
% translation based ones (i.e., TransE) performs better than similarity based  ones (i.e., DistMult) on xxx
% This is because these graphs are sparse, and TransE is less hampered by the sparsity in comparison to the similarity-based techniques \cite{PujaraAG17}.

\subsection{Knowledge Graph Completion}\label{sec:kgc_experiments}

\textbf{Datasets.}
% We evaluate zero-shot knowledge graph completion task on two benchmarks proposed by \cite{Qin2020ZSGAN}, i.e., NELL-ZS and Wikidata-ZS, which are extracted from two well-known knowledge graphs -- NELL\footnote{\url{http://rtw.ml.cmu.edu/rtw/}} and Wikidata\footnote{\url{https://www.wikidata.org/}} respectively.
We evaluate the zero-shot knowledge graph completion task on two benchmarks proposed by \cite{Qin2020ZSGAN}, i.e., NELL-ZS and Wikidata-ZS extracted from NELL\footnote{\url{http://rtw.ml.cmu.edu/rtw/}} and Wikidata\footnote{\url{https://www.wikidata.org/}} respectively, two knowledge graphs are also known for constructing few-shot KGC datasets \cite{chen2019metaR}.
The dataset statistics are listed in Table~\ref{tab:datasets_kgc}.

\textbf{Ontological Schema for KGC.}
Different from the personally defined ontological schemas for IMGC task, many KGs inherently have ontologies which abstractly summarize the entities and relations in knowledge graphs.
Therefore, we access public ontologies of KGs and make a reorganization to construct the ontological schemas we need.
Specifically, for NELL, we process the original ontology file\footnote{\url{http://rtw.ml.cmu.edu/resources/results/08m/NELL.08m.1115.ontology.csv.gz}} and filter out four kinds of properties  to describe the high-level knowledge about NELL relations, i.e., the {\it kgc:domain} and {\it kgc:range} properties which constrain the types of the head entity and the tail entity of a specific relation, respectively, the {\it kgc:generalizations} property which describes the hierarchical structure of relations and entity types, and the {\it kgc:description} property which introduces the literal descriptions of relations and entity types.
While for Wikidata, we utilize Wikidata toolkit packaged in Python\footnote{\url{https://pypi.org/project/Wikidata/}} to access the knowledge of Wikidata relations, in which the {\it kgc:P2302} is used to describe the domain and range constraints of relations, and {\it rdfs:subPropertyOf} and {\it rdfs:subClassOf} are used to describe the hierarchical structure of relation and entity types.
Apart from the textual descriptions of relation and entity types, we also leverage the properties {\it kgc:P31}, {\it kgc:P1629} and {\it kgc:P1855} as the additional knowledge. 
The statistics of the processed ontological schemas are shown in Table\ref{tab:datasets_kgc}.
It is noted that we can also take the original ontologies of these two KGs, but some ontology simplification techniques such as ~\cite{WWTPA2014} may be needed to forget the irrelevant  concepts or properties for prediction tasks contained in the ontologies.
We will consider to develop them in the future.

\begin{table}[]
\small
\caption{Statistics of the zero-shot knowledge graph completion datasets.
\# Ent. and \# Triples denote the number of entities and triples in KGs.
\# Rel. (Tr/V/Te) denotes the number of KG relations for training/validation/testing.
\# Onto. (Trip./Con./Pro.) denotes the number of the RDF triples/concepts/properties in the ontological schemas.
}
  \label{tab:datasets_kgc}
\begin{tabular}{lccc}
\hline
% \toprule[0.6pt]
\multirow{2}{*}{\bf Datasets} 
& \multicolumn{1}{c}{\multirow{2}{*}{\bf \# Ent. \& Triples }} 
& \multicolumn{1}{c}{\bf \# Rel.}
& \multicolumn{1}{c}{\bf \# Onto.} \\
 & \multicolumn{1}{c}{}                          

 & \multicolumn{1}{c}{ Tr/V/Te} 
 & \multicolumn{1}{c}{Trip./Con./Pro.} 
\\\hline
NELL-ZS  
& 65,567 / 188,392  & 139/10/32 
& 3,055/1,186/4  \\
Wikidata-ZS                   
& 605,812 /  724,967 & 469/20/48
& 10,399/3,491/8
\\
\hline
% \bottomrule[0.6pt]
\end{tabular}
\end{table}

\begin{table*}[]
\small
\caption{Results (\%) of zero-shot knowledge graph completion with unseen relations.
The underlined results are the best in the whole column, while the bold results are the best in the pre-training group.
}
\label{tab:kgc_results}
\begin{tabular}{c|l|cccc|cccc}
\hline
% \toprule[0.7pt]
\multirow{2}{*}{\textbf{\begin{tabular}[c]{@{}c@{}}Pre-trained\\ KG Embedding\end{tabular}}}
& \multicolumn{1}{c|}{\multirow{2}{*}{\bf Methods}} 
& \multicolumn{4}{c|}{\bf NELL-ZS}
& \multicolumn{4}{c}{\bf Wikidata-ZS}     \\
                     
& & $MRR$ & $Hit@10$ & $Hit@5$ & $Hit@1$ & $MRR$ & $Hit@10$ & $Hit@5$ & $Hit@1$ 
 \\
\hline
% \midrule[0.5pt] 
\multirow{3}{*}{TransE} 
% & $\text{ZS-TransE}^{\dagger}$ 
& ZS-TransE (Paper) 
& 9.7  & 20.3  & 14.7 & 4.3
& 5.3  & 11.9  & 8.1  & 1.8
\\
& ZSGAN  
& 23.4  & 37.3 & 30.4 & 16.0
& 17.7  & 25.8 & 20.7 & 13.1     \\
& OntoZSL
& {\bf 25.0}  
& {\bf \underline{39.9}} 
& {\bf \underline{32.7}} 
& {\bf 17.2}  
& {\bf 18.4} 
& {\bf 26.5}
& {\bf 21.5}
& {\bf 13.8}  \\
% & OntoZSL-DistMult 
% &     &       &        &           
% &     &         &        &        \\
\hline
%  \midrule[0.5pt] 
\multirow{3}{*}{DistMult} 
% & $\text{ZS-DistMult}^{\dagger}$
& ZS-DistMult (Paper)
& 23.5  & 32.6 & 28.4  & 18.5 
& 18.9  & 23.6 & 21.0  & 16.1        \\
& ZSGAN  
& 24.9  & 37.6 & 30.6  & 18.3
& 20.7  & 28.4 & 23.5  & 16.4   \\
&  OntoZSL
& {\bf \underline{25.6}}  
& {\bf 38.5} & {\bf 31.8}   & {\bf \underline{18.8}}       
& {\bf \underline{21.1}} 
& {\bf \underline{28.9}}
& {\bf \underline{23.8}}
& {\bf \underline{16.7}}  \\
% & OntoZSL     
% &    &         &        &        
% &     &         &        & 
% \\
\hline
% \bottomrule[0.7pt]
\end{tabular}
\end{table*}

\textbf{Baselines and Metrics.}
We mainly compare our proposed OntoZSL with the ZSGAN proposed in \cite{Qin2020ZSGAN}, which generates embeddings for unseen KG relations from their textual descriptions and entity type descriptions.
In ZSGAN and our OntoZSL, the feature extractor can be flexibly incorporated with different pre-trained KG embeddings.
In view of generalization, we adopt two representative KG embedding models TransE \cite{bordes2013translating} and DistMult \cite{yang2014embedding} in the feature extractor in our experiments.
We also compare the original TransE and DistMult in the zero-shot learning setting, i.e., ZS-TransE and ZS-DistMult, where the randomly initialized relation embeddings are replaced by their textual embeddings which are trained together with entity embeddings by the original score functions.
% Besides, we also evaluate two variants of our framework, i.e., {\bf OntoZSL-Trans} and {\bf OntoZSL-Mult}.

% Metrics
As mentioned in Section~\ref{kgc_formulation}, the KGC is to complete (rank) the tail entity given the head entity and relation in a triple.
Therefore, we adopt two metrics commonly used in the KGC literature \cite{wang2017knowledge}: mean reciprocal ranking ($MRR$) and $Hit@k$ to evaluate the prediction results of all testing triples.
$MRR$ represents the average of the reciprocal predicted ranks of all correct entities; while $Hit@k$ denotes the percentage of testing samples whose correct entities are ranked in the top-$k$ positions.
The $k$ is often set to $1, 5, 10$.

\textbf{Implementation.}
We pre-train the feature extractor to extract $200$-dimensional relation embedding for NELL-ZS and extract $100$-dimensional relation embedding for Wikidata-ZS.
In pre-training, the margin parameter $\gamma_f$ is set to $10$, we also follow \cite{Qin2020ZSGAN} to split $30$ triples as references, and set the learning rate to $0.0005$.

We adopt the same ontology encoder configurations and parameters as IMGC to learn $600$-dimensional class embeddings for KGC task.
Especially, the TF-IDF features \cite{salton1988term} are used to evaluate the importance of words in textual descriptions.
Also, we adopt the same GAN architecture as IMGC.
But the difference is, for NELL-ZS, the generator has $400$ hidden units and outputs $200$-dimensional relation embeddings, while the discriminator has $200$ hidden units and outputs a $2$-dimensional vector to indicate whether the input embedding is real or not.
While for Wikidata-ZS, the hidden units of the generator is $200$ and that of the discriminator is $100$.
The dimension of noise vector $z$ is set to $15$ for both datasets, and the number of generated relation embeddings $N_g$ is $20$.
The weight $\lambda_1$ for classification loss is set to $1$, $\lambda_2$ for pivot regularization is set to $3$.
Other parameters for training GAN are identical to those in IMGC task.

% w2c & o2v
\begin{figure*}[htbp]
\centering
\subfigure[Standard ZSL]{
\begin{minipage}[c]{0.5\linewidth}
\centering
\includegraphics[width=6.8cm]{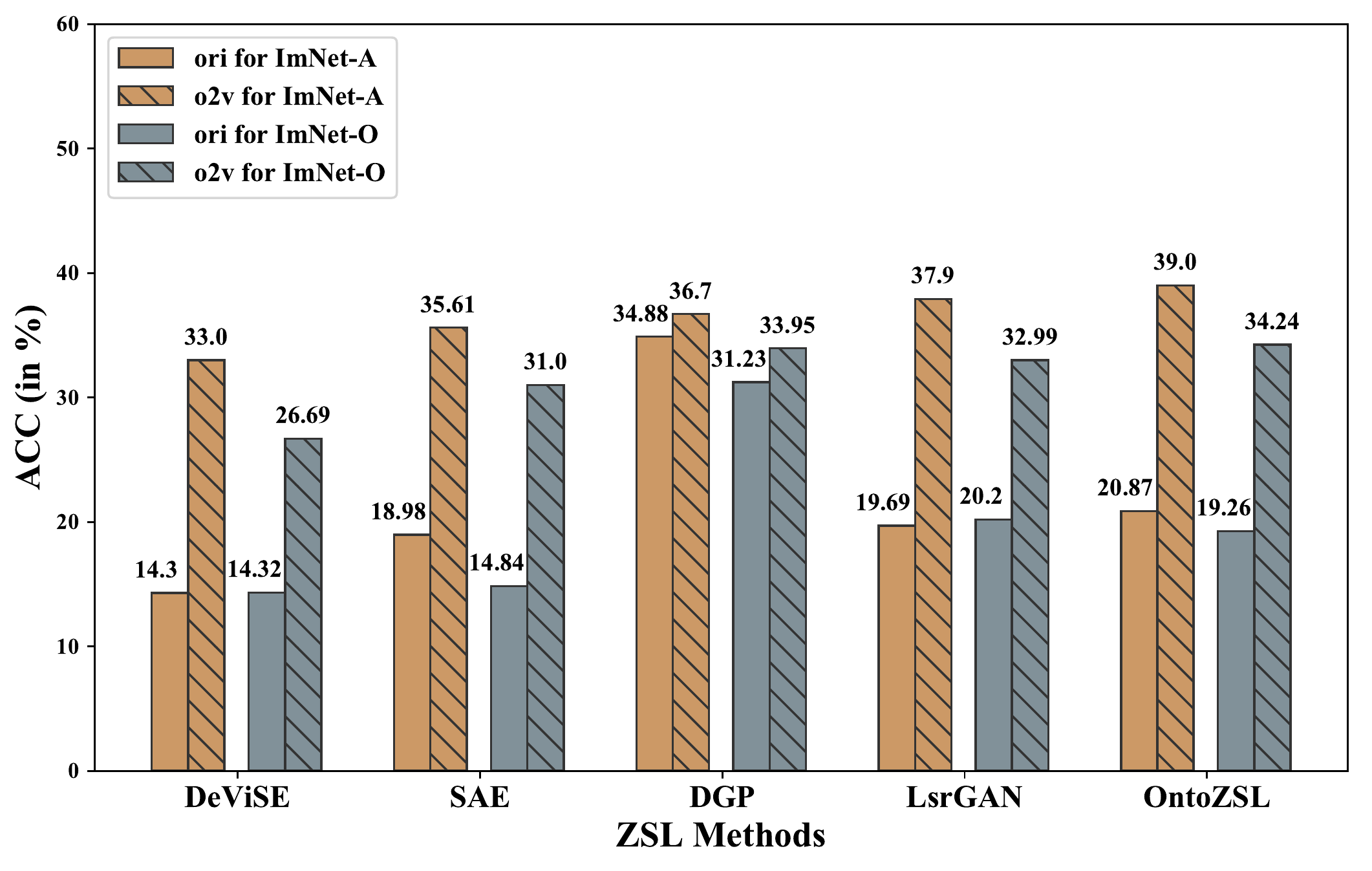}
\end{minipage}%
}%
\subfigure[Generalized ZSL]{
\begin{minipage}[c]{0.5\linewidth}
\centering
\includegraphics[width=6.8cm]{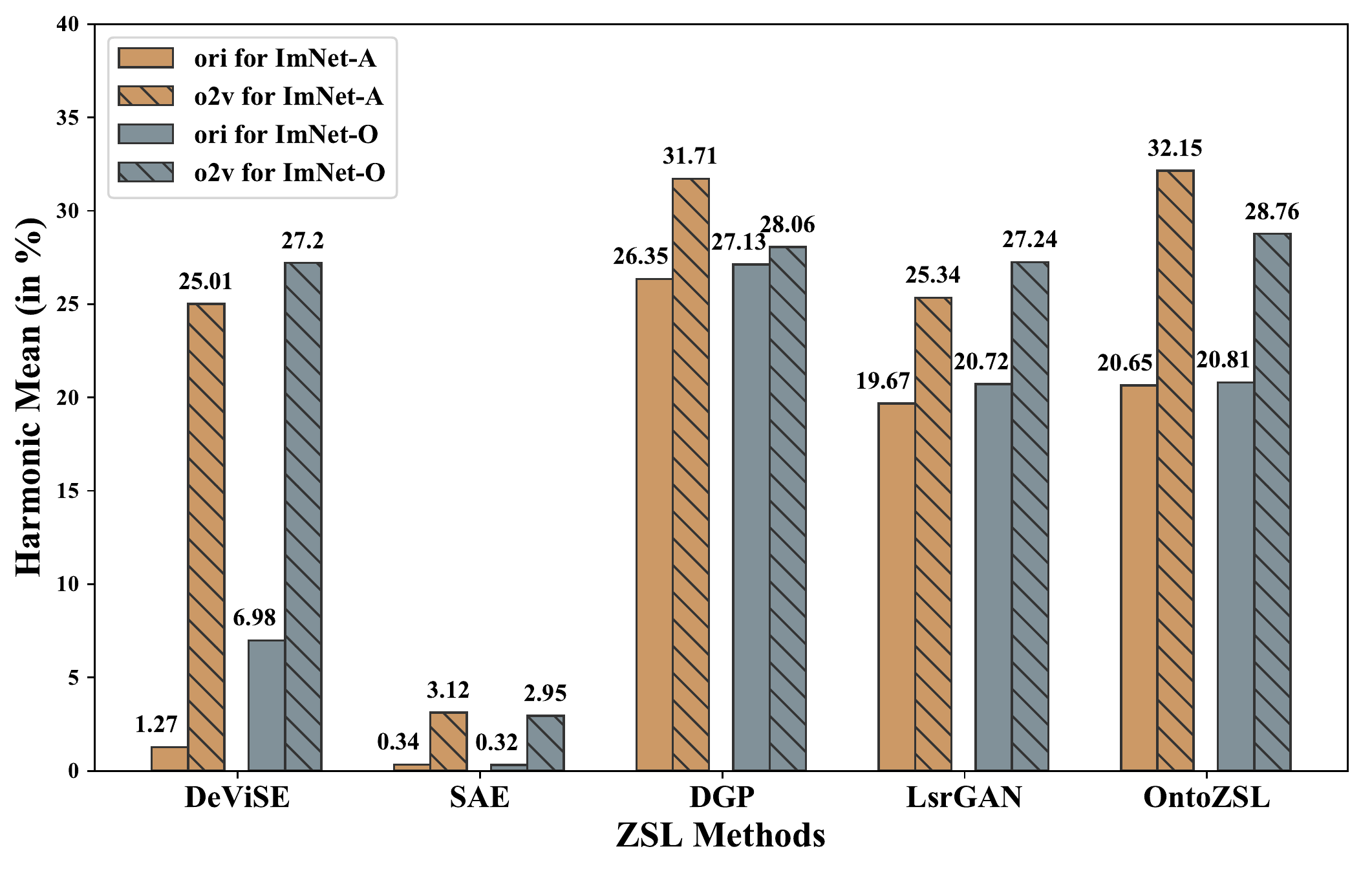}
\end{minipage}%
}%
\centering
\caption{Performance of using different priors w.r.t. different ZSL methods on ImNet-A and ImNet-O.
``ori'' denotes the original class embedding by word2vec or class hierarchy;
``o2v'' denotes the ontology-based class embedding.
ACC (resp. Harmonic Mean) is reported for the standard (resp. generalized) ZSL setting.
}\label{fig:class_embed_results}
\end{figure*}

\textbf{Results.}
Considering that the dataset splits proposed in \cite{Qin2020ZSGAN} are rather new in this domain and the authors do not provide any explanations for the splits,
for a fairer comparison, we conduct experiments with originally proposed train/validation splits as well as with random splits for $3$-fold cross-validation.
Notably, the testing relations are fixed, only the training and validation set are redistributed (i.e., $139$ training relations and $10$ validation relations for NELL-ZS, $469$ training and $20$ validation relations for Wikidata-ZS).
We evaluate our method and ZSGAN on these $4$ splits for both datasets and report the average results in Table~\ref{tab:kgc_results}.
The results of ZS-TransE and ZS-DistMult from original paper, referred as ZS-TransE (Paper) and ZS-DistMult (Paper), are included in the comparison.

In Table~\ref{tab:kgc_results}, we categorize the results into two groups, based on the different pre-trained KG embeddings.
In each group, our OntoZSL achieves consistent improvements over baselines on both datasets.
It indicates that the prior knowledge of KG relations that exists in the ontological schema is superior to that in the textual descriptions.
It is also observed that a higher improvement is achieved when the score function used for the ontology encoder is consistent with that used for pre-training KG embeddings.
For example, compared with ZSGAN on NELL-ZS, the performance is improved by $2.6\%$ on $Hit@10$ with TransE-based pre-trained KG embedding (see the second and third row of Table~\ref{tab:kgc_results}), while is only improved by $0.9\%$ on $Hit@10$ with DistMult-based KG embedding (see the fifth and sixth row of Table~\ref{tab:kgc_results}).

\subsection{Impact of Ontological schema}
To further validate the effectiveness of our ontology-based class semantics, we compare the capabilities of different prior knowledge by applying different class embeddings to multiple ZSL methods including some representative baselines as well as ours.
% In the experiments on image classification datasets ImNet-A and ImNet-O,
Taking the experiments on image classification datasets ImNet-A and ImNet-O as examples,
the originally used word embeddings of classes and the ontology-based class embeddings are applied to the baselines including DeViSE, SAE, DGP and LsrGAN and our method, respectively.
For DeViSE, SAE and LsrGAN, the original class embeddings can be directly replaced with the ontology-based class embeddings we learned, while for DGP which involves word embeddings of classes and class hierarchy, we add attribute nodes produced in our ontological schema into the hierarchical graph to predict the unseen classifiers.

As reported in Figure~\ref{fig:class_embed_results}, the ontology-based class embedding achieves higher performance for all the methods.
For those methods that use class word embeddings as priors, the ontology-based class embeddings have a more than $12\%$ increment on two datasets under the standard ZSL setting, and a more than $2.5\%$ increment under the GZSL setting.
In particular, the harmonic mean of DeViSE increases from $1.27\%$ to $25.01\%$ on ImNet-A and from $6.98\%$ to $27.20\%$ on ImNet-O.
On the other hand, our method expectedly has worse performance after using word embeddings of classes as priors.
Our ontology-based class semantics also improve the performance of DGP when we add attributes to its original class semantics. For example, on ImNet-A, its performance is improved by $1.82\%$ in the standard ZSL setting and by $5.36\%$ in the GZSL setting.
To sum up, our ontology-based class embedding which includes richer class semantics actually performs better than those traditional priors and is beneficial to kinds of ZSL methods.

\subsection{Impact of Ontology Components}
In this subsection, we evaluate the contribution of different components of the ontological schemas by analyzing the performance drop when one of them is removed.
Specifically, with crippled ontological schema, we retrain the ontology encoder to generate class embedding, and then take it as the input of generation model to synthesize unseen features.
We conduct experiments on both two tasks.
For IMGC, we respectively consider removing the literal descriptions, class hierarchy and class attributes in the ontological schema, while for KGC, we consider removing relation constraints (i.e., domain and range constraints), relation and entity type hierarchy, and literal descriptions.
The results on ImNet-A and NELL-ZS are shown in Table~\ref{tab:ablation_imgc} and Table~\ref{tab:ablation_kgc}, respectively. The results on NELL-ZS are based on the KG pre-training setting of TransE.

\begin{table}[]
\small
\caption{Results of OntoZSL on ImNet-A when textual descriptions (``-text''), class hierarchy (``-hie'') or class attributes (``-att'') are removed from the ontological schema.}
\label{tab:ablation_imgc}
\begin{tabular}{l|c|ccc}
\hline
& Standard ZSL &
 \multicolumn{3}{c}{ Generalized ZSL} 
 \\
& $acc$   & \ $acc_s$ \     & \ $acc_u$ \     & \quad $H$ \quad    
\\\hline
% \ \  all     
% & 38.72  & 37.92  & 26.65 & 31.30  \\
% - text      
% & 38.41  & 48.40  & 22.31 & 30.54   \\
% - hierarchy 
% & 35.26  & 51.60  & 20.14 & 28.97   \\
% - attribute 
% & 32.98  & 35.60  & 22.04 & 27.23
\ \  all     
& 39.00  & 37.86  & 27.94 & 32.15  \\
- text      
& 37.63  & 35.07  & 28.50 & 31.45   \\
- hie 
& 35.50  & 39.29  & 24.35 & 30.07   \\
- att 
& 33.88  & 38.07  & 23.71 & 29.22
\\\hline       
\end{tabular}
\end{table}

% 1. in IMGC, compare the attribute and hierarchy
% 2. in KGC, compare the domain&range and hierarchy
% 3. summary: combing the above two components is better; text in two tasks both have contribution. 
From Table~\ref{tab:ablation_imgc}, we can see that the performance of zero-shot image classification is significantly declined when the class attributes are removed under both standard and generalized ZSL settings.
This may be due to the following facts.
First, the attributes describe quite discriminative visual characteristics of classes.
Second, the classes in ImNet-A are fine-grained.
They contain some sibling classes whose differences by the taxonomy are not discriminative.
One example is {\it Horse} and {\it Zebra} in Figure~\ref{fig:example}.
It is hard to distinguish the testing images of such classes when there is a lack of attribute knowledge.
As for the KGC task, as shown in Table~\ref{tab:ablation_kgc}, we find that the hierarchy of relation and entity type has a great influence on the performance.
It is probably because around $58\%$ of the relations in the NELL-ZS dataset are hierarchically related, while only nearly $30\%$ of them have identical domain and range constraints.

\begin{table}[]
\small
\caption{Results of OntoZSL on NELL-ZS when textual descriptions (``-text''), relation and entity type hierarchy (``-hierarchy'') or relation constrains (``-domain\&range'') are removed from the ontological schema.}
\label{tab:ablation_kgc}
\begin{tabular}{l|cccc}
\hline
    & $MRR$ & $Hit@10$ & $Hit@5$ & $Hit@1$ \\\hline
\ \ all    
& 0.250  & 39.9   & 32.7  & 17.2   \\
- text            
& 0.247  & 39.7   & 32.5  & 16.8   \\
- hierarchy  & 0.221  & 35.8 & 29.5  & 14.7   \\
- domain \& range & 0.243  & 38.0  & 31.6 & 16.7  
\\\hline
\end{tabular}
\end{table}
We also find that the performance on both two tasks are slightly influenced when the literal descriptions are removed, indicating the class semantics that exist in text are weak or redundant compared with other semantics.
However, when all of these semantics combined, the best results are achieved.
This means these different components of the ontological schema all have a positive contribution to the ZSL model and are complementary to each other.

\section{Conclusion and Outlook}

In this paper, we propose to use an ontological schema to model the prior knowledge for ZSL.
It not only effectively fuses the existing priors such as class hierarchy, class attributes and textual descriptions for image classes, but also additionally introduces more comprehensive prior information such as relation value constraints for KG relations.
Accordingly, we develop a new ZSL framework OntoZSL, in which a well-suited semantic embedding technique is used for ontology encoder and a Generative Adversarial Network is adopted for feature generation.
It achieves higher performance than the state-of-the-art baselines on various datasets across different tasks.
The proposed ontological schemas are shown to be more effective than the traditional prior knowledge.

In this work, we mainly focus on the newly-added KG relations for the KGC task.
In the future, we would extend our OntoZSL to learn embeddings for those newly-added entities.
Furthermore, we also plan to further extend OntoZSL to explore other tasks such as those related to natural language processing.

\begin{acks}
This work is funded by 2018YFB1402800/NSFCU19B2027/NSFC91846204.
Jiaoyan Chen is mainly supported by the SIRIUS Centre for Scalable Data Access (Research Council of Norway, project 237889) and Samsung Research UK.
\end{acks}

\balance
\bibliographystyle{ACM-Reference-Format}
\bibliography{references}


\appendix
\section{Appendix: Datasets Construction and Attribute Annotation}\label{datasets_attribute_annotation}
In this appendix, we provide more details of %extracting two subsets (i.e., 
ImNet-A and ImNet-O (from ImageNet) and their manually annotating attributes.% for each class in these two datasets.

\subsection{Extracting Classes}
ImageNet  \cite{DBLP:conf/cvpr/DengDSLL009} is a widely used image classification dataset consisting of labeled images of 21 thousand classes.
We conditionally extract different class families from its fine-grained parts.
Classes in a family have the same type, in which 
1) the unseen classes are 1-hop or 2-hops away from the seen classes according to the WordNet texonomy (enabling the transferability from seen classes to unseen classes);
2) the total number of seen and unseen classes is more than $3$ (making the classification is fine-grained);
and 3) each class has a Wikipedia entry (ensuring valid attribute descriptions from Wikipedia for human annotation).

With these conditions, we extract a domain-specific subset ImNet-A, which consists of $11$ animal families, such as {\it Bees}, {\it Ants}, and {\it Foxes}, and a general subset ImNet-O, which consists of $5$ general object families, such as {\it Snack Food} and {\it Fungi}.

\subsection{Preparing Attribute List}
Before annotating attributes, we first prepare the candidate attribute list.
Inspired by the attribute annotations of AwA, which describe the color, shape, texture and important parts of objects, we reuse some attributes from it as well as extract textual phrases which characterize the appearances of classes from Wikipedia entries.
Fox example, one sentence of Wikipedia that describes the class {\it Spoonbill} is: ``Spoonbills are most distinct from the ibises in the shape of their bill, which is long and flat and wider at the end.", from which we can conclude the attribute {\it long flat and wider bill}.

\subsection{Class-Specific Attribute Annotation}
We invite volunteers to manually annotate attributes for these classes.
Specifically, for each class, annotators are asked to assign $3 \sim 6$ attributes from the attribute list with $25$ images given as the reference.
Each class is reviewed by $3 \sim 4$ volunteers, and we take the consensus as the final annotations.
Finally, we annotate a total of $85$ attributes for ImNet-A classes and $40$ attributes for ImNet-O classes.
We associate these attributes with their corresponding classes to construct the ontological schema.









\end{document}